# Hidden Flaws Behind Expert-Level Accuracy of Multimodal GPT-4 Vision in Medicine

Qiao Jin[1], Fangyuan Chen[2], Yiliang Zhou[3], Ziyang Xu[4], Justin M. Cheung[5],

Robert Chen[6], Ronald M. Summers[7], Justin F. Rousseau[8], Peiyun Ni[9],

Marc J Landsman[10], Sally L. Baxter[11], Subhi J. Al'Aref[12], Yijia Li[13],

Alexander Chen[14], Josef A. Brejt[14], Michael F. Chiang[15],

Yifan Peng[3,*] and Zhiyong Lu[1,*]

## Brief Abstract (70 words, following word limit)

We conducted a comprehensive evaluation of GPT-4V's rationales when solving NEJM Image Challenges. We show that GPT-4V achieves comparable results to physicians regarding multi-choice accuracy (81.6% vs. 77.8%). However, we discovered that GPT-4V frequently presents flawed rationales in cases where it makes the correct final choices (35.5%), mostly in image comprehension. As such, our findings emphasize the necessity for in-depth evaluations before integrating such multimodal AI models into clinical workflows.

## Abstract (163 words, preferred)

Recent studies indicate that Generative Pre-trained Transformer 4 with Vision (GPT-4V) outperforms human physicians in medical challenge tasks. However, these evaluations primarily focused on the accuracy of multi-choice questions alone. Our study extends the current scope by conducting a comprehensive analysis of GPT-4V's rationales of image comprehension, recall of medical knowledge, and step-by-step multimodal reasoning when solving *New England Journal of Medicine* (NEJM) Image Challenges – an imaging quiz designed to test the knowledge and diagnostic capabilities of medical professionals. Evaluation results confirmed that GPT-4V performs comparatively to human physicians regarding multi-choice accuracy (81.6% vs. 77.8%). GPT-4V also performs well in cases where physicians incorrectly answer, with over 78% accuracy. However, we discovered that GPT-4V frequently presents flawed rationales in cases where it makes the correct final choices (35.5%), most prominent in image comprehension (27.2%). Regardless of GPT-4V's high accuracy in multi-choice questions, our findings emphasize the necessity for further in-depth evaluations of its rationales before integrating such multimodal AI models into clinical workflows.

**Author affiliations**

[1]National Library of Medicine, National Institutes of Health, Bethesda, MD, USA.

[2]University of Pittsburgh, Pittsburgh, PA, USA.

[3]Department of Population Health Sciences, Weill Cornell Medicine, New York, NY, USA.

[4]Ronald O. Perelman Department of Dermatology, New York University Grossman School of Medicine, New York City, NY, USA.

[5]Department of Medicine, Harvard Medical School and Massachusetts General Hospital, Boston, MA, USA.

[6]Pathology & Laboratory Medicine, Weill Cornell Medicine, New York, NY, USA.

[7]Imaging Biomarkers and Computer-Aided Diagnosis Laboratory, Department of Radiology and Imaging Sciences, National Institutes of Health Clinical Center, Bethesda, MD, USA.

[8]Department of Neurology, Peter O'Donnell Jr. Brain Institute, UT Southwestern Medical Center, Dallas, TX, USA.

[9]Division of Gastroenterology, Department of Medicine, Harvard Medical School and Massachusetts General Hospital, Boston, MA, USA.

[10]Division of Gastroenterology, Department of Medicine, Metrohealth Medical Center, Cleveland, OH, USA. Case Western Reserve University School of Medicine, Cleveland, OH, USA.

[11]Division of Ophthalmology Informatics and Data Science, Viterbi Family Department of Ophthalmology and Shiley Eye Institute, University of California San Diego, La Jolla, CA, USA.

[12]Division of Cardiology, Department of Internal Medicine, University of Arkansas for Medical Sciences, Little Rock, AR, USA.

[13]University of Pittsburgh Medical Center, Pittsburgh, PA, USA.

[14]Department of Internal Medicine, Weill Cornell Medicine, New York, NY, USA.

[15]National Eye Institute, National Institutes of Health, Bethesda, MD, USA.

**\*Corresponding authors**

Yifan Peng, Ph.D., FAMIA

Assistant Professor

Department of Population Health Sciences

Weill Cornell Medicine

425 E 61ST ST DIV 305

New York City, NY 10065

Tel: +1 646-962-9227

E-mail: yip4002@med.cornell.edu

Zhiyong Lu, Ph.D., FACMI, FIAHSI

Senior Investigator

Deputy Director for Literature Search

National Center for Biotechnology Information

National Library of Medicine

National Institutes of Health

8600 Rockville Pike

Bethesda, MD 20894, USA

Tel: 301-594-7089

E-mail: zhiyong.lu@nih.gov

## Main

Large language models (LLMs) exemplified by Generative Pre-trained Transformer 4 (GPT-4)[1] have achieved remarkable performance on various biomedical tasks[2], including summarizing medical evidence[3], assisting in literature search[4,5], answering medical examination questions[6-9], and matching patients to clinical trials[10]. However, most of these LLMs are unimodal, utilizing only the free-text context, while clinical tasks often require the integration of narrative descriptions and multiple types of imaging tests[11,12]. Recently, OpenAI released GPT-4 with Vision (GPT-4V), a state-of-the-art multimodal LLM that allows users to analyze both images and texts together. Subsequent pilot studies have been conducted to analyze the performance of GPT-4V in the medical domain[13-16] (summarized in Supplementary Table 2). These evaluations mainly focused on the accuracy of GPT-4V in answering multi-choice medical questions, and in some cases, GPT-4V outperformed medical students and even physicians in closed-book settings. However, the multi-choice accuracy might not reflect the actual competence of GPT-4V, and there is no guarantee that correct final choices are based on accurate underlying rationales. Therefore, a thorough analysis is imperative to assess whether the decision-making of GPT-4V is based on sound rationales, rather than arbitrary conjecture.

To bridge this gap, we used 207 multiple-choice questions with single correct answers from *New England Journal of Medicine* (NEJM) Image Challenge as this task is non-trivial (see results of a prior foundation model in Methods). Specifically, we concentrated on evaluating the proficiency of GPT-4V in generating both the final answer and the rationales with respect to three capabilities – (1) **Image comprehension**, where the model describes the provided patient image(s); (2) **Recall of medical knowledge**, where the model generates relevant medical knowledge required to solve the question, such as outlining the radiological characteristics associated with each possible choice; and (3) **Step-by-step reasoning**, where the model demonstrates detailed multimodal reasoning to answer the given question, utilizing the generated content from both image comprehension and recall of medical knowledge. These reflect the multimodal nature of the task as well as the commonly evaluated capabilities for text-only LLMs[6,8].

Figure 1 presents the overall design of this study. A senior medical student collected and answered the questions, establishing a student baseline. We then used a specifically designed prompt to ask GPT-4V to generate rationales in separate sections, which facilitates easier localization of the involved capability (described in Online Methods). GPT-4V responses were manually recorded in independent chat sessions. Each question in the dataset was then categorized into a medical specialty and was annotated by one clinician in that field. A multidisciplinary cohort of nine physicians from different specialties was recruited to answer the questions and evaluate the rationales of GPT-4V based on their expertise, with reference to the official correct answers and explanations provided by NEJM Image Challenge. Detailed evaluation guidelines are described in Online Methods.

The evaluation results are shown in Figure 2. Human performance is evaluated in two settings: the closed-book setting (without using external tools such as a literature search engine), and the open-book setting (with external resources) that reflects the actual clinical scenarios. GPT-4V can be considered using a closed-book setting because web browsing is disabled. First, GPT-4V achieved a higher overall accuracy (81.6%, CI: 75.7%-86.7%) than that of physicians (77.8%, CI: 71.5%-83.3%), although the difference is not statistically significant (Figure 2a). GPT-4V also largely outperforms the senior medical student, who achieved an average accuracy of 61.4% (CI: 54.4%-68.0%) under the closed-book setting, representing a human passing score. The best performance under the open-book setting is achieved by human physicians (95.2%, CI: 91.3%-97.7%). Our findings, therefore, align with the previous ones, which show the expert-level performance of GPT-4V in the closed-book setting[15,16]. To investigate performance in relation to question difficulty, we classified the questions into three levels based on the percentage of correct answers chosen by the users from the NEJM website – easy (69 questions), medium (69 questions), and hard (69 questions). Overall performance correlates with question difficulty – almost all respondent groups showed non-inferior performance in easy questions compared to the other levels. Differences between the studied groups are not significant for easy questions. For the medium-level questions, GPT-4V significantly outperforms the medical student in the closed-book setting, but there is no significant difference between the performance of GPT-4V and the human physicians. Interestingly, for hard questions, human physicians with the open-book setting achieved a significantly higher score than GPT-4V.

Figure 2b displays the confusion matrices of GPT-4V and human physicians. Overall, 36 out of 46 (78.3%) questions that physicians failed to answer in the closed-book setting were correctly answered by GPT-4V. Similarly, 7 out of 10 (70.0%) questions incorrectly answered by physicians in the open-book setting were correctly answered by GPT-4V. This suggests that GPT-4V holds potential in decision support for physicians. Such potential utility can be illustrated via Question 96 (Supplementary Data 1), which all human groups answered incorrectly but GPT-4V successfully deduced tongue ulceration as a rare complication in the context of other manifestations of giant cell arteritis. Overall, only 3 out of 207 questions were answered incorrectly by both physicians (open-book) and GPT-4V, indicating a promising synergy between the current tools and GPT-4V. We next evaluated the rationales of GPT-4V in three dimensions – image comprehension, recall of medical knowledge, and step-by-step reasoning (Figure 2c). We found that image comprehension is the most problematic, with more than 25% of cases containing flawed rationales. For example, GPT-4V mistakenly counted the input image containing three CT images, while there are only two provided in Question 12 (Supplementary Data 2). In contrast, medical knowledge recall is the most reliable, with error rates ranging from 11.6% to 13.0%.

Surprisingly, despite overall satisfactory performance, a closer investigation showed that GPT-4V can still be erroneous in one or multiple rationales when the final answer is correct – these mistakes predominantly occur in image comprehension (27.2%), as opposed to knowledge recall (8.9%) and reasoning (12.4%). For instance, in image comprehension of Question 21 (Supplementary Data 3), GPT-4V correctly identified malignant syphilis with multiple evidence, but it failed to recognize that the two skin lesions presenting at different angles actually arise from the same pathology. GPT-4V could also be logically incomplete while guessing right – in Question 95 (Supplementary Data 4), it failed to exclude Argyll Roberson pupil with a sound reason, a condition which also presents with light-near dissociation but has a drastically different etiology. This showed the incompetence of GPT-4V in distinguishing similar manifestations of medical conditions.

Our evaluation has several limitations. First, we studied a set of 207 questions where each question was annotated by one physician of the corresponding specialty, as human examination is costly and time consuming. Additionally, the distribution of NEJM Image Challenge cases might be biased towards certain specialties, such as dermatology. Although we present a more nuanced prompt design to generate structured rationales, alternative strategies such as MedPrompt[9] may be considered with additional resources. Like other similar studies, we use challenge questions with single correct answers. However, clinicians routinely encounter cases where no potential choices are provided and multiple diagnoses are possible. This requires listing rationales for each differential diagnosis with supportive or excluding evidence. In future studies, we plan to also evaluate the rationales of physicians in answering medical questions for comparison.

In summary, we present a comprehensive evaluation of GPT-4V's rationales in multimodal medical tasks. Although GPT-4V demonstrates superior multi-choice accuracy in closed-book settings, physicians remain superior with open-book tools, especially in hard questions. Moreover, among correctly answered questions, GPT-4V may fail to understand or interpret medical scenarios correctly at individual rationales. Our research also identified image comprehension as the greatest challenge for GPT-4V, with an error rate of over 20%, while medical knowledge recall was the most reliable. This suggests that comprehensive evaluations beyond mere multi-choice accuracy are needed before these models can be integrated into clinical practices.

**Methods**

Collecting NEJM Image Challenge

For the collection of NEJM Image Challenges, we assembled 207 most recent questions (between January 11, 2020, and February 29, 2024) along with their ground-truth explanations and answers at https://www.nejm.org/image-challenge. The proportion of correct answers from NEJM users, which varied between 28% and 88%, was employed to indicate question difficulty. Consequently, the challenges were categorized into three difficulty tiers: "easy" for a 56%-88% correct answer rate, "medium" for 44%-56%, and "hard" for 28%-44%. The medical specialty and imaging modality distributions are shown in Supplementary Figure 1.

Prompting GPT-4V

We used the web version of GPT-4V through https://chat.openai.com/ in October 2023 for images 1-100, and in March 2024 for images 101-207. The prompt we used to evaluate GPT-4V is shown below:

    *{image}*

    *{question}*

    *{choices}*

    *Please first describe the image in a section named "Image comprehension".*

    *Then, recall relevant medical knowledge that is useful for answering the question but is not explicitly mentioned in a section named "Recall of medical knowledge".*

*Finally, based on the first two sections, provide your step-by-step reasoning and answer the question in a section named "Step-by-step reasoning".*

*Please be concise.*

Here {*image*}, {*question*}, and {*choices*} represent the actual image, question, and the set of possible answers for each NEJM Image Challenge, respectively.

## BiomedCLIP

To assess the difficulty of the NEJM Image Challenge for vision-language foundation models, we tested the performance of BiomedCLIP[17], a multimodal LLM that is contrastively pre-trained on a dataset of 15 million figure-caption pairs extracted from biomedical literature. We used BiomedCLIP in a zero-shot setting to predict the correct choice for each question. Specifically, let $E_i$ be the pre-trained image encoder and $E_t$ be the pre-trained text encoder. Both the image and the text encoders are accessed from Hugging Face via [microsoft/BiomedCLIP-PubMedBERT_256-vit_base_patch16_224](). Each NEJM Image Challenge contains an image $I$ and five free-text choices $C_1, ..., C_5$. Each choice is prepended with "This is an image of". Concatenating the question and choices leads to sub-optimal results, probably due to the fact that the questions are much longer than the choices. We first generated the embeddings of the image and all choices with their corresponding encoders, and then computed the logit for each choice by its dot product with the image representation:

$$Logit(C_i) = E_i(I)^T E_t(C_i) \in R \tag{1}$$

The choice with the highest logit will be the predicted answer by BiomedCLIP.

$$\hat{i} = Logit(C_i) \qquad (2)$$

Overall, BiomedCLIP achieved the lowest performance of 25.1% accuracy, only slightly better than chance (20%). This suggests the difficulty of the NEJM Image Challenge for vision-language foundation models of smaller sizes.

Annotations on MTurk

The challenge questions are first triaged into nine medical specialties, including dermatology (34 cases), pathology (17 cases), pulmonology (21 cases), gastroenterology (29 cases), neurology (13 cases), ophthalmology (25 cases), cardiology (13 cases), infectious diseases (21 cases), and other internal medicine (34 cases). The triaging is mainly based on the ground-truth answer and explanation of the question. A senior medical student first provided a closed-book answer for each question as the student baseline, then conducted the triaging. For each specialty, a resident or attending physician was recruited to perform a two-stage annotation. In the initial stage, the physicians were asked to answer the questions of their specialty both without (closed-book) and with (open-book) the use of external resources such as internet searches.

In the second stage of annotations, the human physicians review GPT-4V's responses given the ground-truth explanations and answers provided by the NEJM website, evaluating the presence of errors within each segment of GPT-4V's rationale (Image Comprehension, Recall of Medical Knowledge, Step-by-step Reasoning) and the accuracy of GPT-4V's final answers. GPT-4V's rationale for each capability is labeled as either "Correct", "Partially Correct", or "Incorrect". When "Partially Correct" or "Incorrect" are chosen, the physicians are also required to explain the reasons. Evaluation guidelines are shown below:

*Judging the Image Comprehension Rationales*

- *Correct: The rationale accurately identifies all key visual elements relevant to the question. This includes correct identification of any abnormalities present, normal anatomical structures for context, and any relevant clinical signs. The answer demonstrates a thorough understanding of the visual information presented in the image.*

- *Partially Correct: The rationale identifies most of the key visual elements relevant to the question but may miss minor details. There is a general understanding of the image, but the answer lacks completeness or contains minor inaccuracies that do not significantly impact the overall interpretation.*

- *Incorrect: The rationale fails to identify key visual elements relevant to the question, misinterprets the image, or identifies elements that are not present. The answer demonstrates a lack of understanding of the visual information in the image.*

*Judging the Recall of Medical Knowledge Rationales*

- *Correct: The rationale accurately recalls and applies relevant medical knowledge to the question, including underlying pathophysiology, associated clinical features, etc. The answer is comprehensive and demonstrates a deep understanding of the medical concepts pertinent to the question.*

- *Partially Correct: The rationale recalls most of the relevant medical knowledge and applies it to the question but may have minor inaccuracies or omissions. The rationale demonstrates a general understanding of the medical concepts but lacks depth or precision in certain areas.*

- *Incorrect: The rationale fails to recall relevant medical knowledge or applies incorrect information to the question. There is a significant misunderstanding or lack of knowledge of the medical concepts pertinent to the image.*

*Judging the Step-by-Step Reasoning Rationales*

- *Correct: The rationale provides a logical, well-organized explanation of their thought process. Each step is supported by relevant visual clues from the image, medical knowledge, and clinical reasoning. The reasoning process demonstrates clear understanding and application of diagnostic principles.*

- *Partially Correct: The rationale provides a reasoning process that leads to the final diagnosis or conclusion, but the explanation may skip steps, rely on assumptions not clearly supported by the image or medical knowledge, or include minor logical*

> *flaws. The reasoning shows an attempt at systematic thought but lacks clarity or completeness in places.*
> - *Incorrect: The rationale's reasoning process is severely flawed, disorganized, or illogical, leading to an incorrect conclusion. The explanation may lack relevance to the image or question, ignore key information, or demonstrate a misunderstanding of diagnostic reasoning principles.*

Both stages of annotations were conducted on the Amazon Mechanical Turk (MTurk) platform at https://workersandbox.mturk.com/. The annotation interfaces of the first and second stages of annotations are shown in Supplementary Figures 2 and 3, respectively. The full annotation reports are available at https://huggingface.co/datasets/ncbi/Hidden-Flaws-GPT-4V.

To study the potential inter-rater agreement on this task, we have also recruited additional internal medicine physicians to annotate the 34 cases in the category. These annotations are not included in our main analysis. The Cohen Kappa between the two annotators in selecting the closed-book answers is 0.523, suggesting a moderate level of agreement. The Cohen Kappa between the physicians for annotating the rationale correctness is 0.470 for image comprehension, 0.562 for recall of medical knowledge, and 0.698 for step-by-step reasoning. These represent moderate to substantial levels of agreement among the annotators.

## Specialty-level analysis

Supplementary Table 1 shows the specialty-level results. In most specialties, GPT-4V achieved higher (6 out of 9) or comparable (2 out of 9) results than the human physician, and hidden flaws have also been identified by the majority of the physicians (8 out of 9). Interestingly, through this additional specialty-level analysis, we found that the physician accuracy has a very high Spearman rank correlation (91.7%) with the percentage of hidden flaws the physician identified. This indicates that the capability to find flawed rationales might correlate with the capability to answer the original questions.

## Related work

The related studies on evaluating GPT-4V are summarized in Supplementary Table 2 and are compared to this study. Importantly, our evaluation is the only one that includes both quantitative performance with physicians and systematic rationale evaluations for all answers generated by GPT-4V.

## **Acknowledgments**

This research was supported by the NIH Intramural Research Program, National Library of Medicine, Clinical Center, and National Eye Institute. This project was also sponsored by the National Library of Medicine grant R01LM014344.

## **Data Availability**

The hidden flaw annotations are publicly available through Hugging Face datasets at https://huggingface.co/datasets/ncbi/Hidden-Flaws-GPT-4V.

## Code Availability Statement

The BiomedCLIP models are available at https://huggingface.co/microsoft/BiomedCLIP-PubMedBERT_256-vit_base_patch16_224. We accessed GPT-4 Vision through the Web interface (https://chat.openai.com/) and manually collected its responses.

## Competing Interests

The Authors declare no Competing Non-Financial Interests but the following Competing Financial Interests: R.S. receives royalties for patents or software licenses from iCAD, Philips, ScanMed, PingAn, Translation Holdings, and MGB. R.S. received research support from PingAn.

## Author Contributions

Study concepts/study design, **Q.J.**, **Y.P.**, **Z.L.**; manuscript drafting or manuscript revision for important intellectual content, all authors; approval of the final version of the submitted manuscript, all authors; agrees to ensure any questions related to the work are appropriately resolved, all authors; literature research, **Q.J., F.C.**, **Y.P.**, **Z.L.**; experimental studies, **Q.J., F.C., Y.Z.**; human evaluation**, Z.X., J.C., R.C., R.S., J.R., P.N., M.L., S.B., S.A., Y.L., A.C., J.B.**; data interpretation and statistical analysis, **Q.J., F.C., M.C., Y.P., Z.L.**; and manuscript editing, all authors.

**Figure 1: Evaluation Procedure for GPT-4 with Vision (GPT-4V).** This figure illustrates the evaluation workflow for GPT-4V using 207 NEJM Image Challenges. The example instance is adapted from the New England Journal of Medicine, Xiaojing Tang and Lijun Sun, Encapsulating Peritoneal Sclerosis. Copyright © 2024 Massachusetts Medical Society. Reprinted with permission from Massachusetts Medical Society[18]. **a**, A medical student answered all questions and triaged them into specialties. **b**, Nine physicians provided their answers to the questions in their specialty. **c**, GPT-4V is prompted to answer challenge questions with a final choice and structured responses reflecting three specific capabilities. **d**, The physicians then appraised the validity of each component of GPT-4V's responses based on the ground-truth explanations.

**Figure 2. Evaluation results**. **a**, Average multi-choice accuracies achieved by various models and individuals, segmented by question difficulty. **b,** Confusion matrices showing the intersection of errors made by GPT-4V and human physicians. **c**, Bar graphs representing the percentage of GPT-4V's rationales in each capability area as evaluated by human physicians for accuracy. ***: $p<0.001$, n.s.: not significant.